\documentclass[journal]{IEEEtran}
\usepackage{algorithm}
\usepackage{algorithmic}
\usepackage{graphicx}
\usepackage{caption,subcaption}
\usepackage{float}
\usepackage{amsmath}
\usepackage{amssymb}
\usepackage{mathrsfs}
\usepackage{booktabs}
\usepackage{bbm}

\usepackage[pagebackref,breaklinks,colorlinks]{hyperref}
\ifCLASSINFOpdf
\else
\fi
\UseRawInputEncoding

\begin{document}
%
% paper title
% Titles are generally capitalized except for words such as a, an, and, as,
% at, but, by, for, in, nor, of, on, or, the, to and up, which are usually
% not capitalized unless they are the first or last word of the title.
% Linebreaks \\ can be used within to get better formatting as desired.
% Do not put math or special symbols in the title.
\title{Small Object Detection Model with Spatial Laplacian Pyramid Attention and Multi-Scale Features Enhancement in Aerial Images}
%
%
% author names and IEEE memberships
% note positions of commas and nonbreaking spaces ( ~ ) LaTeX will not break
% a structure at a ~ so this keeps an author's name from being broken across
% two lines.
% use \thanks{} to gain access to the first footnote area
% a separate \thanks must be used for each paragraph as LaTeX2e's \thanks
% was not built to handle multiple paragraphs
%

\author{Zhangjian ~Ji,
        Huijia ~Yan, Shaotong Qiao, Kai ~Feng and Wei Wei ~\IEEEmembership{Member,~IEEE,}
        %and Jiye ~Liang,~\IEEEmembership{Senior Member, IEEE}% <-this % stops a space
\thanks{Zhangjian Ji, Huijia Yan, Shaotong Qiao, Kai Feng and Wei Wei are with the Key Laboratory of Computational Intelligence and Chinese Information Processing of Ministry of Education, School of Computer and Information Technology, Shanxi University, Taiyuan 030006, China E-mail: \{jizhangjian@sxu.edu.cn\}}% <-this % stops a space
% <-this % stops a space
%\thanks{Jiye Liang is with the Key Laboratory of Computational Intelligence and Chinese
%Information Processing of Ministry of Education, Shanxi University,
%Taiyuan 030006, Shanxi, China E-mail: \{ ljy@sxu.edu.cn \}}
}

% note the % following the last \IEEEmembership and also \thanks -
% these prevent an unwanted space from occurring between the last author name
% and the end of the author line. i.e., if you had this:
%
% \author{....lastname \thanks{...} \thanks{...} }
%                     ^------------^------------^----Do not want these spaces!
%
% a space would be appended to the last name and could cause every name on that
% line to be shifted left slightly. This is one of those "LaTeX things". For
% instance, "\textbf{A} \textbf{B}" will typeset as "A B" not "AB". To get
% "AB" then you have to do: "\textbf{A}\textbf{B}"
% \thanks is no different in this regard, so shield the last } of each \thanks
% that ends a line with a % and do not let a space in before the next \thanks.
% Spaces after \IEEEmembership other than the last one are OK (and needed) as
% you are supposed to have spaces between the names. For what it is worth,
% this is a minor point as most people would not even notice if the said evil
% space somehow managed to creep in.

% The paper headers
\markboth{Journal of \LaTeX\ Class Files,~Vol.~14, No.~8, November~2025}%
{Shell \MakeLowercase{\textit{et al.}}: Bare Demo of IEEEtran.cls for IEEE Journals}
% The only time the second header will appear is for the odd numbered pages
% after the title page when using the twoside option.
%
% *** Note that you probably will NOT want to include the author's ***
% *** name in the headers of peer review papers.                   ***
% You can use \ifCLASSOPTIONpeerreview for conditional compilation here if
% you desire.

% If you want to put a publisher's ID mark on the page you can do it like
% this:
%\IEEEpubid{0000--0000/00\$00.00~\copyright~2015 IEEE}
% Remember, if you use this you must call \IEEEpubidadjcol in the second
% column for its text to clear the IEEEpubid mark.

% use for special paper notices
%\IEEEspecialpapernotice{(Invited Paper)}

% make the title area
\maketitle

% As a general rule, do not put math, special symbols or citations
% in the abstract or keywords.
%Recently, human pose estimation mainly focuses on how to design a more effective and better deep network structure as human features extractor, and most designed feature extraction networks only introduce the position of each anatomical keypoint to guide their training process. However, we found that some human anatomical keypoints kept their topology invariance, which can help to localize them more accurately when detecting the keypoints on the feature map. But to the best of our knowledge, there is no literature that has specifically studied it. Thus, in this paper, we present a novel 2D human pose estimation method with explicit anatomical keypoints structure constraints, which introduces the topology constraint term that consisting of the differences between the distance and direction of the keypoint-to-keypoint and their groundtruth in the loss object. More importantly, our proposed model can be plugged in the most existing bottom-up or top-down human pose estimation methods and improve their performance. The extensive experiments on the benchmark dataset: COCO keypoint dataset, show that our methods perform favorably against the most existing bottom-up and top-down human pose estimation methods, especially for Lite-HRNet, when our model is plugged into it, its AP scores separately raise by 2.9\% and 3.3\% on COCO val2017 and test-dev2017 datasets.
\begin{abstract}
Detecting objects in aerial images confronts some significant challenges, including small size, dense and non-uniform distribution of objects over high-resolution images, which makes detection inefficient. Thus, in this paper, we proposed a small object detection algorithm based on a Spatial Laplacian Pyramid Attention and Multi-Scale Feature Enhancement in aerial images. Firstly, in order to improve the feature representation of ResNet-50 on small objects, we presented a novel Spatial Laplacian Pyramid Attention (SLPA) module, which is integrated after each stage of ResNet-50 to identify and emphasize important local regions. Secondly, to enhance the model's semantic understanding and features representation, we designed a Multi-Scale Feature Enhancement Module (MSFEM), which is incorporated into the lateral connections of C5 layer for building Feature Pyramid Network (FPN). Finally, the features representation quality of traditional feature pyramid network will be affected because the features are not aligned when the upper and lower layers are fused. In order to handle it, we utilized deformable convolutions to align the features in the fusion processing of the upper and lower levels of the Feature Pyramid Network, which can help enhance the model's ability to detect and recognize small objects. The extensive experimental results on two benchmark datasets: VisDrone and DOTA demonstrate that our improved model performs better for small object detection in aerial images compared to the original algorithm.
\end{abstract}

% Note that keywords are not normally used for peerreview papers.

% For peer review papers, you can put extra information on the cover
% page as needed:
% \ifCLASSOPTIONpeerreview
% \begin{center} \bfseries EDICS Category: 3-BBND \end{center}
% \fi
%
% For peerreview papers, this IEEEtran command inserts a page break and
% creates the second title. It will be ignored for other modes.
\IEEEpeerreviewmaketitle

\section{Introduction}
% The very first letter is a 2 line initial drop letter followed
% by the rest of the first word in caps.
%
% form to use if the first word consists of a single letter:
% \IEEEPARstart{A}{demo} file is ....
%
% form to use if you need the single drop letter followed by
% normal text (unknown if ever used by the IEEE):
% \IEEEPARstart{A}{}demo file is ....
%
% Some journals put the first two words in caps:
% \IEEEPARstart{T}{his demo} file is ....
%
% Here we have the typical use of a "T" for an initial drop letter
% and "HIS" in caps to complete the first word.

\IEEEPARstart{O}{bject} detection is a fundamental task in modern computer vision, aiming to accurately identify and locate specific objects in images or videos. This task plays a crucial role in a wide range of applications, including but not limited to autonomous driving, video surveillance, disaster search and medical imaging analysis. In recently years, with the development of deep learning technology, some typical object detection methods (e.g., FasterRCNN \cite{author5}, YOLO \cite{author22}, SSD\cite{author23}, CornerNet\cite{author29} and RetinaNet) have obtained significant success in nature images (e.g., MS COCO and Pascal VOC). However, these methods can not obtain satisfactory results in aerial images. The main reasons include three aspects: (1) the aerial images usually compose of lots of small scale objects; (2) targets are sparsely and non-uniformly distributed in the whole aerial image;  (3) the number of targets of different categories is grossly unbalanced in aerial images.
%    Object detection typically involves identifying objects across multiple categories and precisely localizing them within images. The presence of objects at various scales in images adds complexity to the detection task. Small objects, in particular, often lack sufficient appearance information, making it difficult to distinguish them from the background or similar objects. Generally, small objects are defined as those smaller than  $32\times32$ pixels. Despite the significant progress made in object detection algorithms driven by deep learning techniques, detecting small objects remains a challenging problem. In real-world applications, factors such as lighting variations, object occlusion, dense object arrangements, and scale changes further complicate the detection of small objects.

Compared with targets in nature images, these challenges mentioned above cause that the existing most object detection methods based on the deep network can not extract effective feature representation for small objects in aerial images because of the features down-sampling in ConvNets. In order to handle it, early works \cite{author34, author58}mainly increase the resolution of feature map by inputting the high-resolution images or reducing the down-sampling rate of network model so as to improve the detection performance of small objects in aerial images. However, simply expanding the resolution of feature maps will impose a huge computational burden, and can not meet the needs of practical applications.

To utilize the higher resolution and reduce the information loss, a popular direction is to crop the original image into small patches, then resize these patches to the size of input image required by the network model and perform object detection on them, such as uniform cropping \cite{author6}, density cropping \cite{author8,author9} and Cascaded Zoom-in (CZ) detector \cite{author7}. The uniform cropping can help improve the detection accuracy of small objects, but the results achieved by it are not optimal because it does not consider the distribution of the objects for cropping, leading to a majority of cropping patches only contain the background or large objects may be cut into two or more different cropping patches. As the objects in aerial images are usually sparsely and non-uniform distributed, an intuitive way to crop the patches is to utilize the additional learnable modules (such as density maps \cite{author9} and cluster proposal networks \cite{author8}) to perform density-based cropping. This usually needs additional components in the network model, often with multiple stages of training. Thus, although the performance of density crop-based methods is better, practitioners still widely employ uniform cropping because of practical simplicity it offers. In order to bridge the gap between research and practice, the CZ detector \cite{author7} can utilize density crops within the training of a standard detector, offering the simplicity of uniform cropping, yet providing the benefits of density crops. Therefore, in this paper, we select the CZ detector as our basic framework. 

In addition, existing studies have shown that the quality of features representation is very crucial to the performance of object detection, and current most object detection methods based on the deep learning are not able to extract the effective features for small objects. Therefore, another feasible direction is to enhance the features representations of small objects by improving the backbone and neck networks of object detection model. For the backbone network, some researchers introduce the channel attention module \cite{author37}, spatial attention module \cite{author60} or convolutional block attention module \cite{author59} to enhance its features representation power. Nonetheless, with the increase of the layer number in the backbone, the downsampling operations make the resolution of features map reduce, which weakens the features representation of small objects. Thus, in order to address this issue, inspirited by the image super-resolution network architecture in \cite{author18}, we proposed a lightweight spatial laplacian pyramid attention module , which can be plugged into each stage of the backbone network (e.g, ResNet50) to heighten the representation power of small objects.  For most object detection models, the Feature Pyramid Network (FPN) is a common network structure used in neck networks and it can generate the different scales feature maps with rich semantic information to detect objects of varying sizes. However, FPN may underperform when detecting small objects. This limitation arises from the loss of fine details of small objects during the top-down multi-scale feature fusion, often caused by downsampling operations such as pooling or large-stride convolutions. As a result, small objects are represented with less precision on feature maps, often appearing blurred and undersized. Additionally, lower-level feature maps in FPN may not provide sufficient resolution to accurately detect small objects, while higher-level feature maps may lack rich semantic information, making it difficult for the model to differentiate between small objects of different categories. To handle this problem, we designed a multi-scale feature enhancement module to augment the features on the C5 layer, under the guidance of adaptive convolution with different dilation rates. Thereby, when the feature maps in FPN are fused from the top-down method, it can effectively enhance the feature representation of small objects. For the sake of clarity, the main contributions of this work can be summarized as follows:
\begin{itemize}
  %\item We first reveal the vital function of the structure constraints among human anatomical keypoints for human pose estimation model, and propose a plug-and-play explicit anatomical keypoints structure constraint model.
  \item After each stage of ResNet-50, we integrate a novel Spatial Laplacian Pyramid Attention (SLPA) module to create an enhanced backbone, which can highlight crucial local regions in the image to boost its features representation power. 
  \item To avoid the information loss from the top-down fusion in FPN, a Multi-Scale Feature Enhancement Module (MSFEM) is incorporated into the lateral connection of C5 layer to build the new feature pyramid network, which can help the model capture the critical detail information in the image so as to enhance its semantic understanding and features representation power.
  \item For the fusing processing of Feature Pyramid network, we utilize the deformation convolutions to align the features of its upper and lower layers, which enables to enhance the model's ability to detect and recognize small objects.
  \item Quantitative experiments and ablation study on two benchmark datasets demonstrate that our improved model performs better for small object detection in aerial images compared to the original algorithm.
\end{itemize}

 The remainder of this paper is organized as follows: Section \ref{sec:rw} reviews the related works, in Section \ref{sec:PM}, we introduce the proposed methodology in details, Section \ref{sec:exp} presents and analyzes the findings from the ablation studies as well as the comparison experiments, and the conclusions in this work are given in Section \ref{sec:con}.

%-------------------------------------------------------------------------------------------------------------------
\section{Related works}\label{sec:rw}

\subsection{Object Detection}
With the rise of deep learning, object detection methods based on deep learning have been dominated in the filed of natural image object detection. According to the pipeline of object detection methods, they can be roughly categorized into two streams: two-stage \cite{author5, author20, author21} and one-stage \cite{author22, author23, author27, author24} object detectors. Two-stage detectors separate object detection into two steps including region generation and region-wise classifier. Earlier two-stage methods (e.g., RCNN \cite{author21}) first generate the candidate regions of the targets, and then extract the features of these RoIs and classify them, which are not end-to-end trainable models. Subsequently, a learnable component known as the Region Proposal Network (RPN) is presented for extracting the candidate regions, giving birth to end-to-end trainable two-stage object detectors \cite{author5,author20}. In contrast, one-stage object detection methods \cite{author22,author24} don't need to extract the RoIs and classify and regress directly from the anchor boxes. They are easy deployment and applicable to real-time object detection. But generally speaking, due to utilize the RoIAlign operation to align the object's features explicitly, two-stage methods tend to be more accurate than one-stages ones and hence used more in aerial detection \cite{author12, author9, author10, author11}. Recently, as deep learning techniques continue to evolve, one-stage detectors have narrowed the performance gap with two-stage ones. Thus, they are being exolored in aerial images \cite{author13, author61}. Nowadays, anchor-free detectors \cite{author26, author27, author28, author29} have become very popular, since they avoid to hand-craft anchor box dimensions and spend extra time matching them. Thus, in \cite{author13}, the authors also attempt to apply their detection framework on anchor-free FCOS \cite{author27}. In this paper, for a fair comparison with existing methods, we mainly perform our study on the two-stage detectors.
\subsection{Small Object Detection in aerial images}
In addition to natural images, object detection has been widely applied in aerial images. However, most of existing ordinary object detection methods can not achieve the desired detection performance because aerial images contain lots of small scale objects that are sparsely and non-uniformly distributed. In order to handle the inadequacy of information problem for small object detection, many researchers have undertaken extensive work in this field. According to different technical routes, these works can be categorized into: scale-aware training \cite{author16, author53, author13,author36, author30}, super-resolution methods \cite{author34, author58, author62}, context-modeling \cite{author19, author35, author32, author63}, focus-and-detection \cite{author7, author8, author9, author10, author12, author54, author64, author65} and so on. Due to the sparse and non-uniform distribution of small objects in the hight-resolution aerial images, for small object detection in aerial image, density cropping is a popular method, such as ClusDet \cite{author8}, DMNet \cite{author9}, CDMNet \cite{author10} and so on. These methods can guarantee that small objects are processed at higher resolutions, thereby reducing the information loss and improving the performance of subsequent object detection. However, they generally need to training the additional focus-detecting model to generate the candidate regions, which are more complex to use than uniform cropping. Different from those, The CZDet \cite{author7} re-purposes the detector itself to extract the density cropping regions, thereby eliminating the need for an additional learnable module. 

%\subsection{Other methods}
\subsection{Attention Mechanisms}
Attention mechanisms that mimic the human visual system play a critical role in deep neural networks, helping the network focus on the regions of interest that are beneficial for decision-making in images. This mechanism has been successfully applied in various tasks, such as object detection \cite{author67} and semantic segmentation \cite{author66}. For instance, the early attention models, e.g, SENet \cite{author37}, it introduces channel-wise attention module that calculated via global average-pool operation to exploit the inter-channel relationship. However, it only obtains the suboptimal features because of missing the spatial attention that plays an important role in deciding 'where' to focus. The CBAM \cite{author59},  which is a representative model of mixture attention mechanisms, composes of spatial and channel-wise attention modules and is superior to the SENet \cite{author37} that only using the channel-wise attention. But these two models lack the pow of global context modeling effectively enough, the GCNet, combing the optimal implementation of each step of simplified non-local block \cite{author38} and SE block \cite{author37}, can effectively model the global context, with the lightweight property of SENet \cite{author37}. Thus, like SE block, GCNet can be also applied in all residual blocks of the ResNet architecture. Although these attention modules mentioned above achieve certain performance gains on many visual tasks, they are not specifically designed for object detection. To improve the performance of object detection, Li \emph{et al.} \cite{author67} propose a MAD unit to aggressively search for neuron activations among feature maps from both low-level and high-level streams. In addition, most of existing object detection methods only rely on recognizing object instances individually, without exploiting their relations during learning. In fact, besides the contextual information, the relation between objects can also help object detection and recognition, so Hu \emph{et al.} \cite{author46} propose an adapted attension module to model the relations between objects for object detection. Unfortunately, for small object detection in aerial images, these attention models mentioned above do not perform well because downsampling operations make small objects miss lots of useful information. Thus, in order to handle this issue, inspirited by the image super-resolution network architecture in \cite{author18}, we design a lightweight spatial laplacian  pyramid attention module, which can be embedded into each stage of the backbone (e.g, ResNet50) to enhance the features representation power of small objects. 
\subsection{Multi-Scale Feature Enhancement}
Multi-scale feature enhancement addresses the issue of varying object scales by acquiring and fusing features at multiple scales, thereby improving the final detection results. For example, in order to boost the performance of object detection, Lin \emph{et al.} \cite{author16} first proposed a feature pyramid network (FPN) for feature representation,  which combines the low-level location features in shallow layers with high-level semantic features in deep layers. Subsequently, in order to further improve the feature presentation power of FPN, many researchers make every effort to ameliorate the structure of FPN by adding the additional pathways to further fuse deep and shallow features or incorporating attention mechanisms to guide the fusion of features across different layers. E.g., AC-FPN \cite{author51} utilizes dilated convolutions of different rates to capture semantic information from the highest feature layers by fully integrating attention guidance, and PANet \cite{author47} adds an additional bottom-up pathway on top of FPN, which achieves better performance than FPN, but with the cost of more parameters and computations. To improve the efficiency of the model, Bi-FPN \cite{author48} introduces weighted bi-directional cross-scale connections based on PANet to balance information across different scales. TridentNet \cite{author53} leverages dilated convolutions to build parallel multi-branch structures with different receptive fields, handling object scale variation in detection tasks through scale-aware training. AugFPN \cite{author15} adopts a consistency supervision approach to narrow the semantic gaps between different scale features before multi-scale features fusion, reducing information loss in the highest level of FPN. Although those methods mentioned above boost the accuracy of object detection in natural images, their limitations arise when it is expected to capture the densely tiny object in aerial images. Thus, in this paper, under the guidance of adaptive convolution with different dilation rates, we design a multi-scale feature enhancement module to build a new feature pyramid network, which can help the model capture the critical detail information of images to facilitate small objects detecting.

%NAS-FPN \cite{author50} employs neural architecture search to find the better topology of cross-scale connections for FPN. FPT achieves non-local interaction of features across space and scale using attention mechanisms, enabling full fusion of contextual features.
%-------------------------------------------------------------------------
%\section{Proposed human anatomical keypoints constraints model}\label{sec:PM}
\section{Proposed Method}\label{sec:PM}
%In this section, we specifically describes the proposed anatomical keypoints constraints model and how it can be plugged into the existing top-down and bottom-up human pose estimation methods.
In this section, we mainly describe our two basic modules that designed specifically to improve the detection performance for small objects in detail and how it can be plugged into existing small object detection methods.
%\subsection{Generate the coordination set of human anatomical keypoints}\label{sec:l1cft}
\subsection{Improved Object Detection Framework}\label{sec:l1cft}
\begin{figure*}[!h]
  \centering
  % Requires \usepackage{graphicx}com
 \includegraphics[width=14cm]{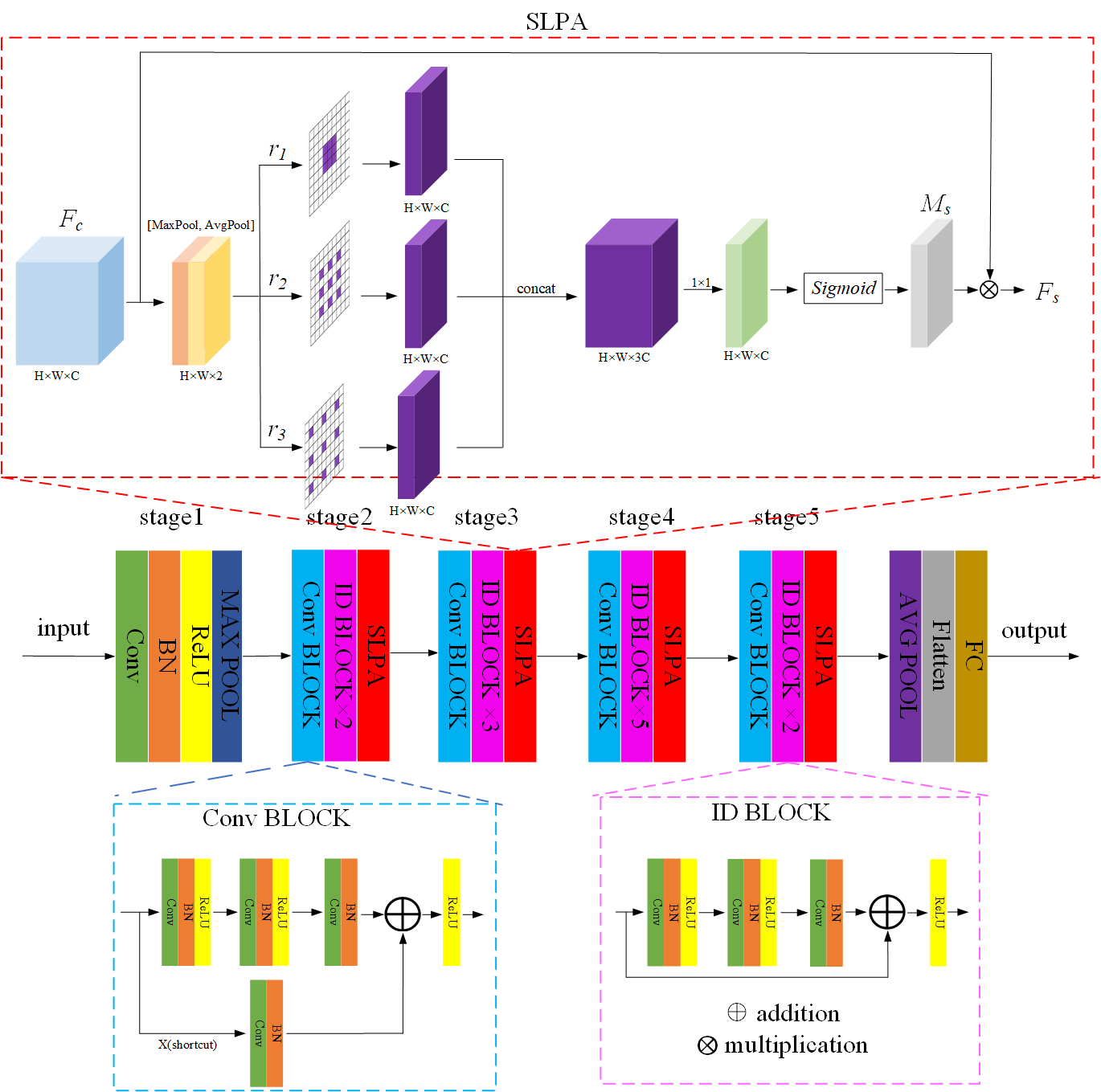}\\
  \caption{The improved ResNet50 architecture after inserting our proposed SLPA module.}\label{fig_1}
\end{figure*}
\begin{figure*}
  \centering
  % Requires \usepackage{graphicx}com
 \includegraphics[width=14cm]{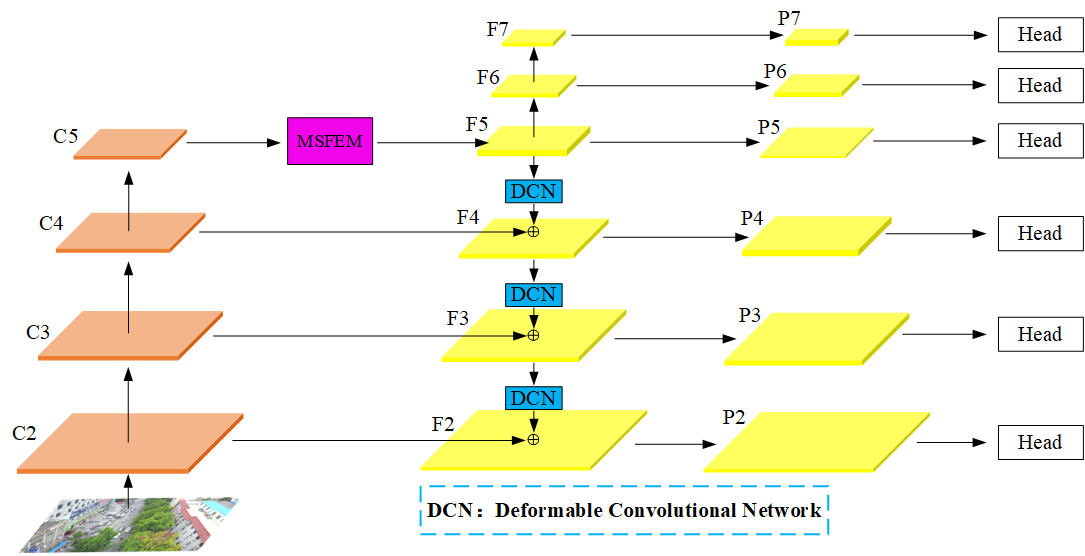}\\
  \caption{The architecture of the improved FasterRCNN-FPN.}\label{fig_2}
\end{figure*}

The FasterRCNN-FPN, as a two-stage object detector, has been widely used in lots of aerial image object detection models, \emph{e.g.},  ClusDet \cite{author8}, DMNet \cite{author9}, CDMNet \cite{author10}, CZDet \cite{author7} and so on. However, due to containing plenty of small or even tiny objects in aerial images, the backbone network (e.g., ResNet) and neck network (e.g., FPN) that adopted in FasterRCNN are not able to extract the effective features for them. Therefore, a feasible direction is to enhance its ability to detect small objects by modifying the architecture of backbone and neck networks in FasterRCNN. For the backbone network, inspirited by the image super-resolution network architecture in \cite{author18}, we designed a lightweight Spatial Laplacian Pyramid Attention (SLPA) module, which can be plugged into most backbone networks to boost their features representation power for small objects. Taking the ResNet50 as an example, our proposed SLPA module is embedded into its each stage, that is, the improved ResNet50 is shown in Figure \ref{fig_1}. For the neck network, the FPN is a common network structure and it can generate the different scales features maps with rich semantic information to detect the objects of varying sizes. However, when perform the top-down feature fusion in FPN, the top-layer feature suffers from some information loss because its channel number decreases, which severely impairs its feature representation capability. Therefore, in order to enhance its feature representation power for small objects, we design a multi-scale feature enhancement module to insert the lateral connection of $C_{5}$ layer of FPN, which extracts and fuses the diverse information of the top-layer feature. Sequently, we improve its architecture  by replacing the FPN of FasterRCNN-FPN with our new FPN module, which is shown in Figure \ref{fig_2}.
\subsection{Spatial Laplacian Pyramid Attention Module}
With the increase of layer number in the backbone (e.g., ResNet), the ever-shrinking features map weakens the features representation for small objects. In order to enhance the features representation of small objects and capture more subtle features, inspirited by the image super-resolution network architecture in \cite{author18}, we designed a lightweight Spatial Laplacian Pyramid Attention (SLPA) module, which is shown in Figure \ref{fig_1}.

As is shown in the SLPA module of Figure \ref{fig_1}, firstly, the input features $F_{c}\in\mathbb{R}^{H\times W\times C}$ is compressed from C-dimensional channels to 2-dimensional channels via the MaxPool and AvgPool, which can be repressed as
\begin{equation}\label{eq1}
 F^{'}=concat(MaxPool(F_{c}),AvgPool(F_{c})),
\end{equation}
where $concat(\cdot,\cdot)$ denotes the merging of two feature tensors.

Next, to capture the contextual information at the Laplacian Pyramids, we use the ReLU and the different dilation rates convolutional layers to learn the critical features at different scales as follow:
\begin{equation}\label{eq2}
f_{i}=ReLU(r_{i}(F^{'})), (i=1,2,3) ,
\end{equation}
where $r_{i}$ denotes the dilation convolution and its dilation rate is specified by the subscripts. The multi-level features $f_{1}, f_{2}, f_{3}$ obtained from $F^{'}$ are concatenated as $Z=[f_{1};f_{2};f_{3}]$.

Furthermore, to produce the attention map, $Z$ is fed into $1\times1$ convolution operation $Conv_{1\times1}$ followed by the sigmoid activation as:
\begin{equation}\label{eq3}
M_{s}=Sigmoid(Conv_{1\times1}(Z)).
\end{equation}

Finally, the input $F_{c}$ of the Spatial Laplacian Pyramid Attention (SLPA) is adaptively rescaled by the attention map $M_{s} $ as:
 \begin{equation}\label{eq4}
F_{s}=F_{c}\times M_{s}.
\end{equation}
\subsection{Multi-scale Features Enhancement Module}
 Top-down feature fusion in FPN is performed by using the $C_{5}$ layer feature. However, the $C_{5}$ layer feature only contains the semantic information for the specific receptive field but is incompatible with other feature maps. Thus,  inspired by adaptive convolution kernels \cite{author69}, we consider using them to perform the multi-scale feature enhancement for the $C_{5}$ layer feature. For this purpose, we design the module shown in the following Figure \ref{fig_3}. Firstly, in our proposed module, in order to perform multi-scale feature enhancement, we split the $Y$ feature channels output by the $C_{5}$ layer into four groups and use adaptive convolution to convolve each group with different dilation rates \cite{author68}. Meanwhile, the global information of the $C_{5}$ layer is also obtained by an adaptive global average pooling. Next, to ensure that the sufficient feature information is contained, we concatenate the original features $Y$, grouping feature after processed by adaptive convolution and the global information that obtained by an adaptive global average pooling. Finally, we perform a $1\times1$ convolution to allow better integration of feature information. Thus, our proposed multi-scale feature enhance module can be formulated as follows:
\begin{equation}\label{eq5}
y_{i}=\left\{
\begin{aligned}
Y,~~~~~~~~~~&i=0\\
\sigma(Adconv_{r}(split(Y))), ~&i,r\in\{1,2,3,4\}\\
U_{s}(AAP2d(Y)),~&i=5
\end{aligned}
\right.
\end{equation}
 \begin{equation}\label{eq6}
output=Conv_{1\times1}(concat([y_{0},y_{1},y_{2},y_{3},y_{4},y_{5}]))
\end{equation}
where $split(\cdot)$ represents a grouping function, which divides the $Y$ feature into four groups along the channel; $\sigma(\cdot)$ denotes the Leaky ReLU activation function; $AAP2d(\cdot)$ represents the adaptive average pooling operation; $U_{s}$ denotes the upsampling operation; $Adcon_{r}(\cdot)$ defines the adaptive dilation convolution with different dilation rates.
 \begin{figure*}
  \centering
  % Requires \usepackage{graphicx}com
 \includegraphics[width=14cm]{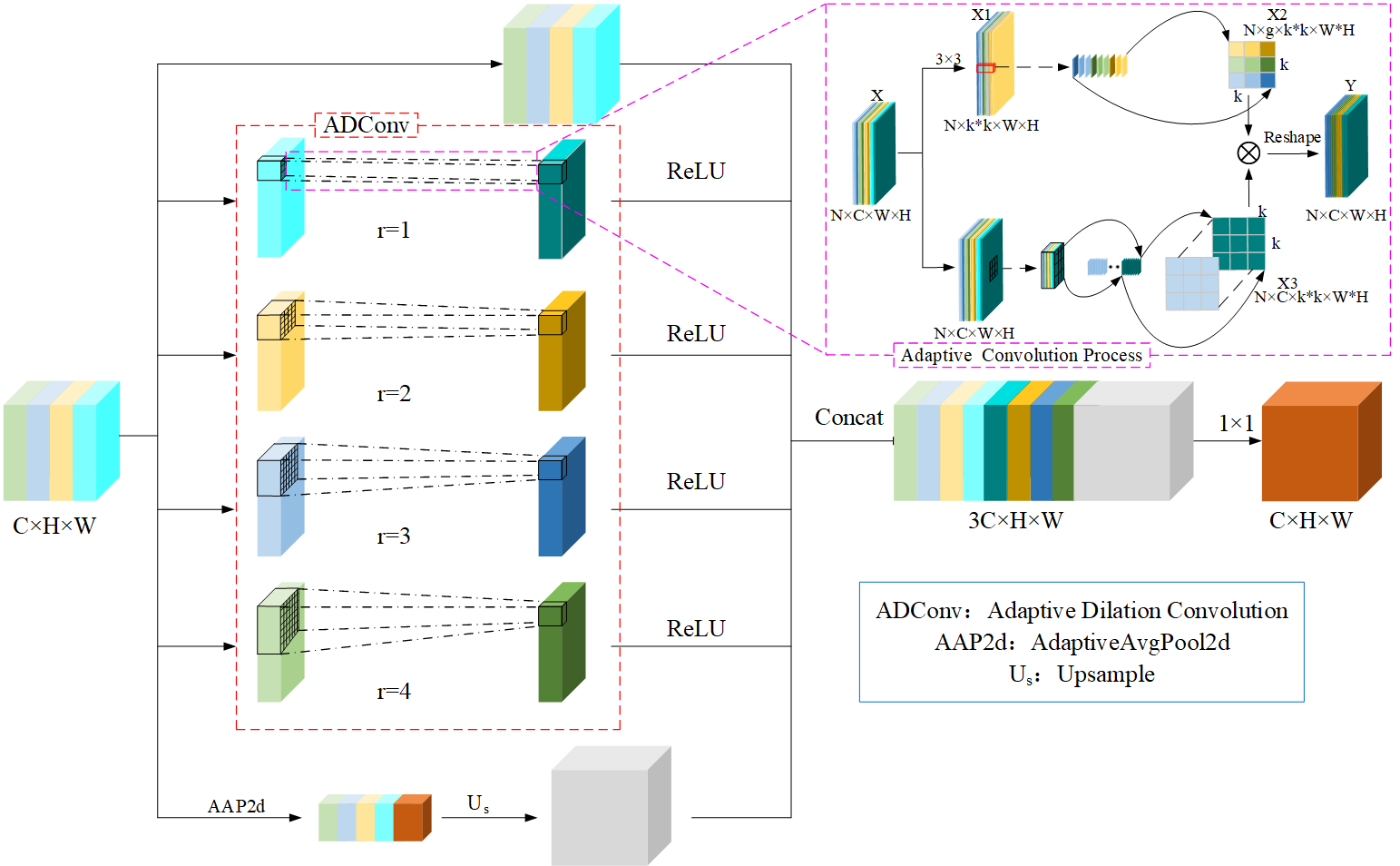}\\
  \caption{The whole process of multi-scale feature enhancement module.}\label{fig_3}
\end{figure*}
 
The adaptive dilation convolution module adopts the dual-branch structure to learn the contextual information around each pixel (seeing the adaptive convolution process of the Figure \ref{fig_3}). In the upper branch, we compress the input feature $X\in\mathbb{R}^{N\times C\times W\times H}$ to the size of adaptive convolution kernel ($X1\in\mathbb{R}^{N\times k\ast k \times W\times H}$, $k$ denotes the kernel size), which obtains the surrounding contextual information on each pixel. After that, we go through a batch normalization layer and a softmax activation layer to obtain the features with surrounding contextual information and take out the feature pixel by pixel on $k\times k$ channels to expand into $k\ast k$ convolution kernel size. Finally, we reshape the dimensions of features to obtain the adaptive convolution kernel parameters $X2\in\mathbb{R}^{N\times g\times k\ast k\times H\ast W}$ ($g$ denotes the number of groups of convolution kernel channels), which can be formulated as: 
\begin{equation}\label{eq7}
X2=reshape(Softmax(Bn(Conv_{3\times3}(X1)))).
\end{equation}
In Eq.\ref{eq7}, the $Conv_{3\times3}$ represents the convolution operation with a convolution size of 3. The $Bn$ is batch normalization. 

Next, in the lower branch, the input feature $X\in\mathbb{R}^{N\times C\times W\times H}$ is expanded pixel by pixel to $k\ast k$ size, obtaining $X3\in\mathbb{R}^{N\times C\times k\ast k\times W\ast H}$. Finally, we adopt the adaptive convolution kernel parameters $X2\in\mathbb{R}^{N\times g\times k\ast k\times H\ast W}$ obtained by the Eq.\ref{eq7} to perform convolution on the same spatial location of $X3$, $i.e.$
\begin{equation}\label{eq8}
Y_{i,j}=reshape(\sum_{p\in\Omega}X2_{p, i*W+j}\cdot X3_{p,i*W+j}),
\end{equation}
where $Y_{i,j}$ denotes the output features of convolution at pixel location $(i,j)$, and $p\in\Omega:=\{1,2,\cdots,k*k\}$ denotes the pixel index within the range of convolution.
%-------------------------------------------------------------------------
\section{Experiments}\label{sec:exp}
%\section{Experiments} %\label(experiment}
In this section, we first introduce the benchmark datasets and evaluation metrics used in our experiments. Next, we describe the implementation details. Then, we compare the proposed SLPA module and the existing CBAM, and conduct ablation study to analyze the effectiveness of the proposed two modules (SLPA and MSFEM) and the effect of different dilation rates in them. Finally, we compare the improved CZ Det that integrates the proposed SLPA and MSFEM with the original CZ Det and other existing state-of-the-art and the most related object detection methods on VisDrone and DOTA-v1.0 datasets. 
%%-------------------------------------------------------------------------
\subsection{Experimental setup}
\noindent\textbf{Datasets.} To assess the performance of our proposed model, we conduct the experimental evaluations on two popular challenging benchmark datasets for aerial image object detection, namely VisDrone \cite{author3} and DOTA \cite{author4} datasets. The VisDrone consists of 10209 aerial images (6471 for training, 548 for validation and 3190 for testing) with a pixel size of about $2000\times 1500$ pixels, the training set of which is manually annotated with 540k instance bounding boxes that are from ten categories of objects. The VisDrone dataset has some problems such as some object being too small, extreme category and scale imbalance and severe target occlusion, which makes it an ideal benchmark for studying small object detection problems. In addition, since the evaluation server is closed now, we can only follow the existing works \cite{author7, author8, author9, author10} to use the validation set for evaluation performance. The DOTA dateset is composed of satellite images, including 1411 training images and 458 validation images. It contains roughly 280k manually annotated instances being from the fifteen different categories of objects, such as planes, ships, large vehicles, helicopters, etc.

\noindent\textbf{Evaluation metric.} Following the standard evaluation protocol that proposed in MS COCO \cite{author70}, we report six evaluation metrics, namely $AP$ , $AP_{50}$, $AP_{75}$, $AP_{s}$, $AP_{m}$ and $AP_{l}$, to measure the performance of different models. Specifically, $AP$ is the average precision under multiple IoU thresholds, ranging from 0.5 to 0.95 with a step size of 0.05. The $AP_{s}$, $AP_{m}$ and $AP_{l}$ are respectively average precision for small (less than $32\times 32$ pixel size), medium and large (larger than $96\times 96$ pixel size) objects, which are suitable to study the performance of different methods for small object detection.
%-------------------------------------------------------------------------
\subsection{Implementation details}
In this paper, all the experiments of our proposed method are run on a single NVIDIA A100 GPU with 40G RAM and implemented based on Detectron2 toolkit\footnote{\url{https://github.com/facebookresearch/detectron2}}. For our experimental validation, we use Faster RCNN as the baseline object detector and its backbone adopts the ResNet50 with FPN that pre-trained on ImageNet dataset. In this paper, our proposed two new modules are plug-and-play type and they can be applied to most existing object detectors, such as CZ Det \cite{author7}, ClusterNet \cite{author8}, DensityMap \cite{author9}, etc. Thus, in order to more fully verify the effectiveness of our proposed new modules in the experiments, we inserted our proposed new modules into three object detection models mentioned above and compared them with the original baseline models. For data augmentation and hyper-parameters, we adopt the same setup with the original baseline models. For VisDrone and DOTA datasets, the model was respectively trained for 180k and 100k iterations. The initial learning rate is both set to 0.007071 and respectively decayed by 10 at 100k and 140k iterations and at the 40k-th and 70k-th epoch.
\subsection{Comparison between our SLPA and the existing CBAM}
In order to verify the effectiveness of our proposed SLPA module, Table \ref{tab:com-1} gives a comparison between our SLPA module and the existing CBAM (it is a mixture attention model, composing of spatial and channel-wise attention modules) on the VisDrone dataset. Here, we select the CZ Det \cite{author7} as our baseline and its object detector (Faster RCNN) adopts the ResNet50 with FPN as the backbone. Baseline+CBAM represents that the CBAM is plugged into each stage of ResNet50 and Baseline+SLPA indicated that our proposed SLPA module is integrated into each stage of ResNet50. Seeing from the Table \ref{tab:com-1}, we find that the performance of our proposed SLPA module is superior to that of the CABM, especially for small and middle objects. The possible reason is that our proposed SLPA that employed a pyramid mechanism induced by convolutions with different expansion rates, is better at capturing the fine-grained features of small and medium-sized targets.
\begin{table*}[!h]
  \caption{Comparison between our proposed SLPA module and the CBAM on the VisDrone dataset. The best results are highlighted by the bold.}
  \centering
  %\resizebox{1.8\columnwidth}{!}{
  \begin{tabular}{cccccccc}
    \toprule
     methods&  Input size & AP(\%) &$AP_{50}$(\%)&$AP_{75}$(\%)&$AP_{s}$(\%)&$AP_{m}$(\%)&$AP_{l}$(\%)\\
    \midrule
    Baseline &$1200\times1999$ &33.2 &58.3&33.2 & 26.1 &42.6&43.4\\
    Baseline+CBAM &$1200\times1999$ &33.9 &59.0&34.0 & 26.6 &43.5&\textbf{44.6}\\
    Baseline+SLPA &$1200\times1999$ &\textbf{34.3} &\textbf{59.5} &\textbf{34.8} &\textbf{26.9} &\textbf{43.9}&44.4\\
    \bottomrule
  \end{tabular}
  %}
  \label{tab:com-1}
\end{table*}
\subsection{Ablation study}
In this subsection, we select the CZ Net as the baseline and design a number of ablation studies to analyze the contribution of our proposed modules from different aspects. All the experiments are performed on the VisDrone dateset and the resolution of the input image is $1200\times1999$.

\noindent\textbf{Effect of each module.} In Table \ref{tab:ablation-1}, we compare the impact of incorporating our proposed modules into the baseline model on its performance. We observe that the AP score of the baseline raises by 1.1\% after only integrating our proposed SLPA module into it. Notably for small and medium-sized objects, the AP scores increase by 1.0\% and 1.3\%, respectively, surpassing the 0.8\% improvement observed in large objects. This result suggests that our proposed SLPA module can improve the features representation ability of the backbone network by the Laplacian Pyramid attention mechanism, especially for small and medium-sized object. When only adding the MSFEM into the baseline model, its performance is improved by 1.3 AP, which shows that it is feasible to utilize our proposed MSFEM to enhance the top-level feature map, because it helps the model mine the critical detail information in the image. In addition, seeing from Table \ref{tab:ablation-1}, we also find that the AP Scores of small, medium and large-sized objects have all demonstrated improvements across different magnitudes. Especially for large object, its AP score improves by 1.1\%, outperforming the improvements in the small and medium-sized objects. The main reason is that the top-level feature is inherently responsible for detecting large objects. As shown in the Table \ref{tab:ablation-1}, we observe that integrating deformable convolution modules between the layers of the pyramid structure for feature alignment results in a 1.5-percentage-point improvement in the AP score of baseline model. This result, on the other hand, also demonstrates that during the fusion of features across different levels of FPN, the feature misalignment induced by upsampling operations can significantly degrade the model performance. Meanwhile, we also notice that the AP scores of small and medium-sized objects respectively increase by 1.1\% and 1.7\% but the one of large objects drops by 1.8\%. The possible reason is that because the number of small, medium and large-sized objects in VisDrone dataset is unbalance, the offsets that learned by the deformation convolution are biased toward adapting to small and medium-sized objects rather than the large ones, which leads to a decline in the detection performance of large objects. After integrating the LSPA and MSFEM into the baseline, the AP score of the baseline increases by 1.5\%, exceeding the performance of only adding either of two modules, because these two modules work together to provide the model with high-quality feature information (LSPA can enhance local detailed information and MSFEM can enrich multi-scale features).  When integrating the LSPA and DCN modules into the baseline model simultaneously, we observe that the AP score of the baseline model increases by 1.9\%, and the ones of small, medium and large-sized objects are all improved, which is superior to the results obtained by adding either of the two modules separately. The main reason is that combining two modules makes full use of the local details representation ability of SLPA and the DCN's feature alignment in the fusing process. After adding three modules into the baseline simultaneously, the AP score reaches 35.3\%, and other evaluation metrics also achieve the best results. These results fully validate that the mutual cooperation among the three modules can comprehensively enhance the performance of the model.
\begin{table*}
  \caption{Ablation study of three different modules on the VisDrone dataset, conducted by successively incorporating them into the baseline. The best results are highlighted by the bold.}
  \centering
  %\resizebox{1.8\columnwidth}{!}{
  \begin{tabular}{ccccccccc}
    \toprule
     SLPA& MSFEM & DCN & AP(\%) &$AP_{50}$(\%)&$AP_{75}$(\%)&$AP_{s}$(\%)&$AP_{m}$(\%)&$AP_{l}$(\%)\\
    \midrule
     &  &  &33.2& 58.3&	33.2& 26.1&42.6& 43.4\\
     $\surd$ &  & &34.3 &59.5 &34.8 &27.1 &43.9&44.2\\
     &$\surd$ & &34.5 &59.7 &35.0 &26.9 &43.5&44.5\\
      & &$\surd$ &34.7 &60.1 &35.3 &27.2 &44.3&41.6\\
     $\surd$&$\surd$ & &34.7 &60.3 &34.5 &27.0 &44.0&44.9\\
     $\surd$& &$\surd$ &35.1 &60.6 &35.9 &27.8 &45.0&44.6\\ 
     &$\surd$ &$\surd$ &34.8 &59.7 &35.0 &26.9 &44.5&45.0\\  
    $\surd$ &$\surd$ &$\surd$&\textbf{35.3}&\textbf{61.1} &\textbf{36.1} &\textbf{28.0} &\textbf{45.6}&\textbf{45.5}\\
    \bottomrule
  \end{tabular}
  %}
  \label{tab:ablation-1}
\end{table*}

\noindent\textbf{Effect of different dilation rates in our SLPA and MSFEM modules.} In Table \ref{tab:ablation-2}, we give the results of different dilation rate configurations in our SLPA module. Seeing the Table \ref{tab:ablation-2},  we note that the performance of SLPA module performs better when its dilation rate configuration is set to $\{1,2,3\}$, especially for small object, its AP score reaches the peak. For other alternative dilation rate configurations, the overall performance of the SLPA module exhibits fluctuation, with certain key indicators showing a decline (e.g., $AP_{s}$), which illustrates that different dilation rate configurations can influence the ability of the SLPA module to capture the detailed features at different scales, and an appropriate dilation rate configuration can enhance its ability to learn the contextual information in the image better, thereby boosting the overall performance of the proposed model. Thus, in our experiments, the dilation rate configuration of the SLPA module is set to $\{1,2,3\}$. 

In addition, Table \ref{tab:ablation-3} shows the results of different dilation rate configurations in our MSFEM. Seeing from it, we also observe that when the dilation rate configuration of the MSFEM is $\{1,2,3,4\}$, its performance performs best.  If selecting other dilation rate configuration, the overall performance of the MSFEM has declined to a certain extent, which verifies that different dilation rate configurations are crucial to the multi-scale feature enhancement of the MSFEM and affect its ability to capture and fuse the different scale features. So, we set the dilation rate configuration of the MSFEN to $\{1,2,3,4\}$ in our experiments.
\begin{table*}
  \caption{Ablation study of different dilatation rate configurations within our SLPA module. The best results are highlighted by the bold.}
  \centering
  %\resizebox{1.995\columnwidth}{!}{
  \begin{tabular}{ccccccc}
    \toprule
     Dilatation rates (r) & AP(\%) &$AP_{50}$(\%)&$AP_{75}$(\%)&$AP_{s}$(\%)&$AP_{m}$(\%)&$AP_{l}$(\%)\\
    \midrule
    \{1,2,3\}&\textbf{34.3} &\textbf{59.5} &\textbf{34.8}&\textbf{27.1}&43.9 &44.4\\
    \{1,2,4\}&34.2 &59.1 &\textbf{34.8} &26.8 &43.7 &44.1\\
    \{1,2,5\}&\textbf{34.3}&59.0&\textbf{34.8}&27.0&	\textbf{44.2}&	44.2 \\
    \{1,3,5\}&34.0&	58.9&	34.6&	26.7&	43.8&	43.5\\
    \{1,3,6\}&33.9&	58.6&	34.6&	26.5&	44.1&	\textbf{44.5}\\
    \{1,3,9\}&34.0&	58.9&	34.7&	26.7&	43.8&	44.1 \\
    \bottomrule
  \end{tabular}
  %}
  \label{tab:ablation-2}
\end{table*}
\begin{table*}
  \caption{Ablation study of different dilatation rate configurations within our MSFEM module. The best results are highlighted by the bold.}
  \centering
  %\resizebox{1.995\columnwidth}{!}{
  \begin{tabular}{ccccccc}
    \toprule
     Dilatation rates (r) & AP(\%) &$AP_{50}$(\%)&$AP_{75}$(\%)&$AP_{s}$(\%)&$AP_{m}$(\%)&$AP_{l}$(\%)\\
    \midrule
    \{1,2,3,4\}&\textbf{34.5}& \textbf{59.7}&	\textbf{35.0}&	\textbf{26.9}&	43.5&	\textbf{44.5}\\
    \{1,2,3,5\}&34.3&	59.6&	34.2&	26.1&	43.2&	42.6\\
    \{1,2,4,6\}&34.4&	\textbf{59.7}&	34.5&	26.5&	\textbf{44.1}&	42.7 \\
    \{1,3,5,7\}&34.1&	59.5&	34.1&	26.0&	43.2&	42.3\\
    \bottomrule
  \end{tabular}
  %}
  \label{tab:ablation-3}
\end{table*}
%\begin{figure}[!h]
%  \centering
%  % Requires \usepackage{graphicx}
%  \includegraphics[width=7.5cm,height=6cm]{Figures/AP_curve.png}\\
%  \caption{Curve of the AP scores of three different methods (Lite-HRNet, Lite-HRNet+Length and Lite-HRNet+Length+Angle) on COCO $val2017$ dataset in the training process. Here, Lite-HRNet+Length denotes that only length constraint term of our proposed keypoints constraint model is added into the Lite-HRNet and Lite-HRNet+Length+Angle represents that the Lite-HRNet contains our human keypoints constraint model defined in the formula (\ref{eq3}).}\label{fig_7}
%\end{figure}
%
\begin{table*}[!h]
  \caption{Comparison results on the VisDrone2019 dataset. The symbol * denotes that the corresponding object detection methods use the improved backbone network, that is, our proposed SLPA module is plugged into each stage of ResNet-50. The symbol $\dag$ indicates that the results of the corresponding object detection methods after their input image size is adjusted from $1000\times1999$  to $600\times1000$. `MF' stands for model fusion. The best result of one column is highlighted by the bold.}
  \centering
  \resizebox{1.995\columnwidth}{!}{
  \begin{tabular}{ccccccccc}
    \toprule
     Methods & Backbone & Input size & AP(\%) &$AP_{50}$(\%)&$AP_{75}$(\%)&$AP_{s}$(\%)&$AP_{m}$(\%)&$AP_{l}$(\%)\\
     \midrule
     \multicolumn{9}{c}{One-stage methods }\\
    \midrule
     QueryNet \cite{author13}&ResNet-50&-&28.3 &48.1 &28.8 &19.8 &35.9&40.3\\
     SDPNet \cite{author55}&ResNet-50&-&33.7 &56.6 &34.3 &26.7 &42.9&45.7\\
     Uniform crop\cite{author6}&ResNet-50&$1024\times1024$&31.7 &56.3 &31.6 &25.1 &40.4&41.1\\
     LMFF-MFEE \cite{author71}&YOLOv8s&$640\times640$&27.2 &44.5&- &- &-&-\\
     \midrule
     \multicolumn{9}{c}{Two-stage methods}\\
     \midrule
     ClusterNet \cite{author8}&ResNet-50 &$600\times1000$ & 26.7&50.6 &24.7 &17.6 &38.9&51.4\\
     ClusterNet* &ResNet-50+SLPA &$600\times1000$ & 26.9&50.7 &24.7 &17.8 &39.0&51.5\\
     DensityMap \cite{author9}&ResNet-50 &$600\times1000$ & 28.2&47.6 &28.9 &19.9 &39.6&55.8\\
     DensityMap*&ResNet-50+SLPA &$600\times1000$ & 28.5&47.9 &29.1 &20.1 &39.8&55.9\\
     CDMNet \cite{author10}&ResNet-50 &$600\times1000$ & 29.2&49.5 &29.8 &20.8 &40.7&41.6\\
     AMRNet \cite{author11}&ResNet-50 &$800\times1500$ & 31.7&- &- &23.0 &43.4&58.1\\
     CRENet \cite{author72}&Hourglass-104 &$1024\times 1024$ & 33.7&54.3 &33.5 &25.6 &45.3&\textbf{58.7}\\
     UCGNet \cite{author54}&DarkNet-53 &- & 32.8&53.1 &33.9 &- &-&-\\
     GLSAN \cite{author12}&ResNet-101 &$600\times1000$ & 30.7&55.6 &29.9 &- &-&-\\
     CascadeNet \cite{author14}&ResNet-50 &-& 28.8&47.1 &29.3 &- &-&-\\
     CascadeNet+MF \cite{author14}&ResNet-50 &-& 30.1&58.0 &27.5 &- &-&-\\
     CZ Det\dag &ResNet-50&$600\times1000$ &28.5&51.5&28.0&20.8&38.1&50.7\\
     CZ Det*\dag &ResNet-50+SLPA&$600\times1000$ &29.2&52.0&28.5&21.6&38.4&51.1\\
     CZ Det \cite{author7}&ResNet-50&$1000\times1999$ &33.2&58.3&33.2&26.1&42.6&43.4\\
     CZ Det* &ResNet-50+SLPA&$1000\times1999$ &34.3&59.5&34.8&26.9&43.9&44.4\\
     CZ Det+ours &ResNet-50+SLPA&$1000\times1999$ &\textbf{35.3}&\textbf{61.1}&\textbf{36.1}&\textbf{\textbf{28.0}}&\textbf{45.6}&45.5\\
    \bottomrule
  \end{tabular}
  }
  \label{tab:res1}
\end{table*}

\begin{table}[!h]
  \caption{Complexity comparison between the CZ Det and the improved CZ Det model with our proposed modules. }
  \centering
  \resizebox{0.995\columnwidth}{!}{
  \begin{tabular}{cccccc}
    \toprule
     Methods & Backbone & FLOPs/G & Params/M &FPS&AP\\
     \midrule
     CZ Det\cite{author7}&ResNet-50&213.12&100.7&12.0&33.2\\
     CZ Det+ours&ResNet-50+SLPA&218.22&107.8&11.4&35.3\\
    \bottomrule
  \end{tabular}
  }
  \label{tab:para}
\end{table}

\begin{table*}
  \caption{Comparison results of the different object detection algorithms on DOTA-v1.0 dataset. The best result of one column is highlighted by the bold.}
  \centering
  \resizebox{1.995\columnwidth}{!}{
  \begin{tabular}{cccccccccc}
    \toprule
     Methods & Backbone & Input size & AP(\%) &$AP_{50}$(\%)&$AP_{75}$(\%)&$AP_{s}$(\%)&$AP_{m}$(\%)&$AP_{l}$(\%)&FPS\\
     \midrule 
     Uniform crop\cite{author6}&ResNet-50 &$1024\times1024$ & 33.4&54.0 &35.6 &16.9 &36.8&43.7&\textbf{0.49}\\
     ClusterNet \cite{author8}&ResNet-50 &$1000\times1000$ & 32.2&47.6&\textbf{39.2}&16.6&32.0&\textbf{50.0}&-\\
     CZ Det\cite{author7}&ResNet-50 &$1200\times1999$ & 34.6&	56.9&	36.2&	18.2&	37.8&	43.8&0.30\\
     CZ Det+ours&ResNet-50+SLPA &$1200\times1999$ &\textbf{35.0}&	\textbf{57.6}&	36.8&	\textbf{20.2}&	\textbf{38.3}&	43.6&0.24\\
    \bottomrule
  \end{tabular}
  }
  \label{tab:test}
\end{table*}
%%-------------------------------------------------------------------------
\subsection{Results}
In this subsection, to comprehensively validate the effectiveness of our proposed SLPA and MSFEM modules, we mainly select the CZ Det as the baseline, and compare its performance after adding our proposed modules with other existing state-of-the-art and the most related methods, including one-stage (QueryNet \cite{author13}, SDPDet \cite{author55}, Uniform Crop \cite{author6} and LMFF-MFEE \cite{author71}) and two-stage (ClusterNet \cite{author8}, DensityMap \cite{author9}, CDMNet \cite{author10}, AMRNet \cite{author11}, GLSAN \cite{author12}, CascadeNet \cite{author14}, CZ Det \cite{author7} and so on) small object detection methods, on VisDrone2019 and DOTA-v1.0 datasets. 

\noindent\textbf{VisDrone2019.} Table \ref{tab:res1} gives the comparison results on VisDrone2019 dataset. We observe that, under the same input resolution ($600\times1000$), the performance of CZ Det is superior to that of some similar methods (e.g., ClusterNet and Density Map), especially for small target, with the AP score increasing by 3.2\% and 0.9\% respectively. After adopting the improved ResNet50 backbone, namely integrating the proposed SLPA module into each stage of ResNet50, the performance of three model above is improved to a certain extent, specifically for CZ Det, its AP and $AP_{s}$ scores respectively raise by 0.7\% and 0.8\%. This result illustrates that the proposed SLPA module is helpful for improving the performance of object detector because it enhances the feature representation of the backbone (ResNet50), making the model capture the key details in the image. As shown in Table \ref{tab:res1}, the performance of CZ Det is significantly improved when the input resolution follows the setting of $1200\times1999$ as specified in the original CZ Det paper \cite{author7}, with AP and $AP_{s}$ increasing by 4.7\% and 5.3\%, respectively,  compared to those obtained using a $600\times1000$ input resolution. Moreover, there results surpass the AP and $AP_{s}$ of CZ Det variant that adopts the improved backbone (ResNet-50+SLPA) and $600\times1000$ input resolution by 4.0\% and 4.5\%, respectively. The main reason is that the larger input resolution can provide more richer detailed information for detecting the targets, especially the small ones, which is far more useful than using any feature enhancement method. Meanwhile, we also notice that when the CZ Det with $1200\times1999$ input resolution adopts the improved backbone (ResNet-50+SLPA) to be operated, the AP, $AP_{s}$, $AP_{m}$ and $AP_{l}$ scores exceed the corresponding results reported in the baseline by 1.1\%, 0.8\%, 1.3\% and 1.0\%, respectively. If continuing to integrate our proposed MSFEM module and the existing DCN into the CZ Det, there results can be further improved by 1.0\%, 1.1\%, 1.7\% and 1.1\%.

Table \ref{tab:para} reports the complexity comparison between the CZ Det and the improved CZ Det model with our proposed modules. Compared to the CZ Det, after it adds our proposed modules, the computational load Flops and the number of parameters have slightly increased, and the detection frame rate FPS only drops from 12.0 to 11.4, which is entirely within the acceptable range. The results show that without sacrificing computational efficiency excessively, the proposed modules significantly improve object detection accuracy.

\noindent\textbf{DOTA-v1.0.} To further verify the effectiveness of the proposed modules, we also compare several different algorithms on DOTA-v1.0, as shown in Table \ref{tab:test}. The variant of CZ Det, namely integrating the proposed modules, its AP score reaches 35.0\%, which has an improvement compare to the baseline model (CZ Det), especially for small targets, the $AP_{s}$ score increases from 18.2\% to 20.2\%. The results illustrates that the proposed modules help the baseline improve the detection performance of small targets in aerial images, and better adapt to the variations of target size and shape in remote sensing images. For large objects, the ClusterNet performs better than our model. The reason is probably that our model gets biased to detect small objects, thanks to the additional crops on training.
\begin{figure*}
  \centering
  % Requires \usepackage{graphicx}
  \includegraphics[width=17.5cm,height=6cm]{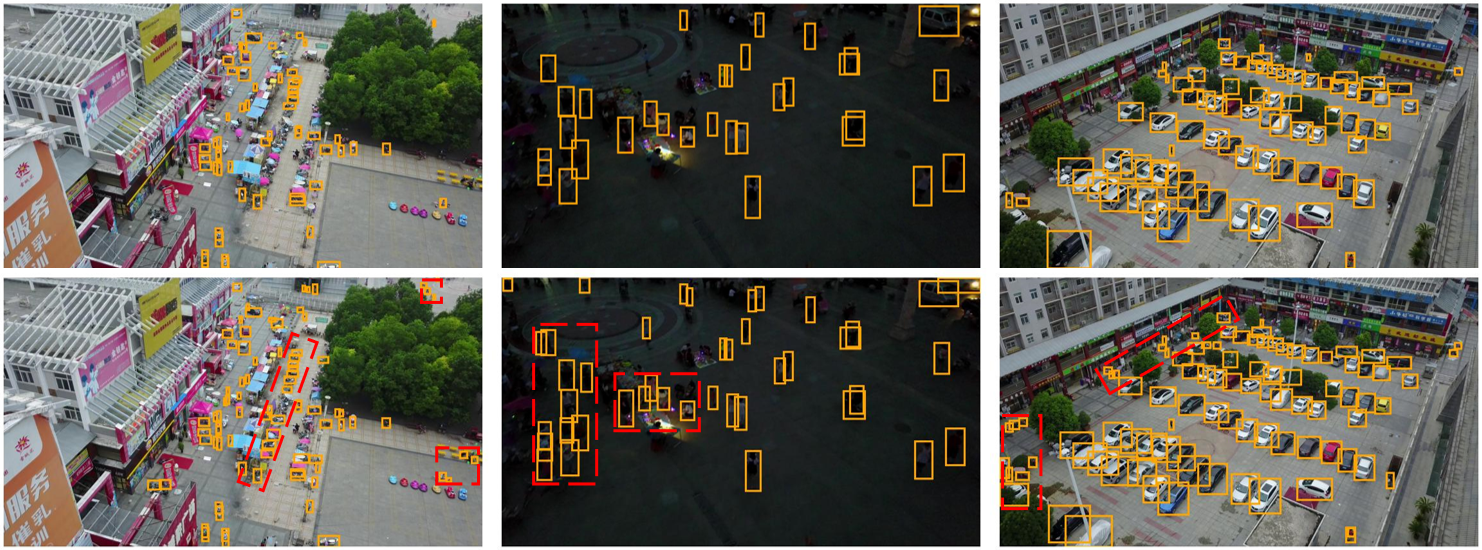}\\
  \caption{Visualization results of the original CZ Det (the first row) and the improved CZ Det (the second row) on several representative scene images.}\label{fig_vis}
\end{figure*}

\noindent\textbf{Visualization.} In order to clearly and intuitively compare the object detection results of the original CZ Det and the improved CZ Det that integrates our proposed models in this paper, in Figure \ref{fig_vis}, we show their detection results on several representative scenarios, including high-density crowded and low-light nighttime situations, where the yellow box represents the detected targets while the red dotted-line box contains the new targets detected by the improved CZ Det besides the original detection results. It can be seen from the visualization results that in various scenarios, the improved CZ Det performs the detection performance better. Especially in the severe occlusion and low-light situations, such as nighttime scenes, the improved CZ Det can detect the more targets and reduce the rate of missing detection effectively. The main reason is that our proposed modules can help the model extract more discriminative features of the targets.

\section{Conclusion}\label{sec:con}

This paper first analyzes the challenges faced in small object detection and the shortcomings of feature pyramids in this context. We then introduce the SA attention mechanism after each stage of ResNet-50 to enhance the model's perceptual and representational capabilities, thereby improving small object detection performance. Additionally, we integrate the MSFEM module at the deepest layer of the feature pyramid to enhance the model's understanding of semantic information and improve feature transfer across different levels, thereby strengthening feature representation. We also add DCN modules between each layer of the feature pyramid to enhance multi-scale feature fusion and feature alignment. Finally, experimental results demonstrate that the improved algorithm achieves superior performance on datasets such as VisDrone, enhancing the original algorithm's effectiveness in small object detection.

% if have a single appendix:
%\appendix[Proof of the Zonklar Equations]
% or
%\appendix  % for no appendix heading
% do not use \section anymore after \appendix, only \section*
% is possibly needed

% use appendices with more than one appendix
% then use \section to start each appendix
% you must declare a \section before using any
% \subsection or using \label (\appendices by itself
% starts a section numbered zero.)
%

%\appendices
%\section{Proof of the First Zonklar Equation}
%Appendix one text goes here.
%
%% you can choose not to have a title for an appendix
%% if you want by leaving the argument blank
%\section{}
%Appendix two text goes here.

% use section* for acknowledgment
\section*{Acknowledgment}
This work is supported by the National Natural Science Foundation of China (No.61602288, 62676160), Fundamental Research Program of Shanxi Province (No.202503021211073). The authors also would like to thank the anonymous reviewers for their valuable suggestions.

% Can use something like this to put references on a page
% by themselves when using endfloat and the captionsoff option.
\ifCLASSOPTIONcaptionsoff
  \newpage
\fi

% trigger a \newpage just before the given reference
% number - used to balance the columns on the last page
% adjust value as needed - may need to be readjusted if
% the document is modified later
%\IEEEtriggeratref{8}
% The "triggered" command can be changed if desired:
%\IEEEtriggercmd{\enlargethispage{-5in}}

% references section

% can use a bibliography generated by BibTeX as a .bbl file
% BibTeX documentation can be easily obtained at:
% http://mirror.ctan.org/biblio/bibtex/contrib/doc/
% The IEEEtran BibTeX style support page is at:
% http://www.michaelshell.org/tex/ieeetran/bibtex/
%\bibliographystyle{IEEEtran}
% argument is your BibTeX string definitions and bibliography database(s)
%\bibliography{IEEEabrv,../bib/paper}
%
% <OR> manually copy in the resultant .bbl file
% set second argument of \begin to the number of references
% (used to reserve space for the reference number labels box)
%%%%%%%%% REFERENCES
{\small
\bibliographystyle{IEEEtran}
\bibliography{referenceBib}
}
%\begin{thebibliography}{1}
%
%\bibitem{IEEEhowto:kopka}
%H.~Kopka and P.~W. Daly, \emph{A Guide to \LaTeX}, 3rd~ed.\hskip 1em plus
%  0.5em minus 0.4em\relax Harlow, England: Addison-Wesley, 1999.
%
%\end{thebibliography}

% biography section
%
% If you have an EPS/PDF photo (graphicx package needed) extra braces are
% needed around the contents of the optional argument to biography to prevent
% the LaTeX parser from getting confused when it sees the complicated
% \includegraphics command within an optional argument. (You could create
% your own custom macro containing the \includegraphics command to make things
% simpler here.)
%\begin{IEEEbiography}[{\includegraphics[width=1in,height=1.25in,clip,keepaspectratio]{mshell}}]{Michael Shell}
% or if you just want to reserve a space for a photo:
 
\begin{IEEEbiography}[{\includegraphics[width=1in,height=1.25in,clip,keepaspectratio]{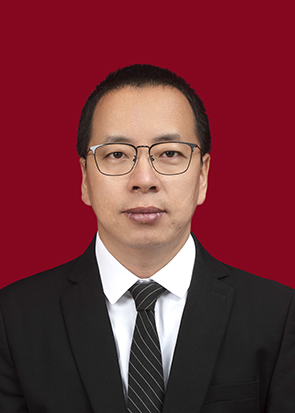}}]{Zhangjian Ji} received the B.S. degree from Wuhan University, China, in 2007, the M.E. degrees from the Institute of Geodesy and Geophysics, Chinese Academy of Sciences (CAS), China, in 2010, and the Ph.D. degree in the University of Chinese Academy of Sciences, China, in 2015.

He is currently an Associate Professor at the School of Computer and Information Technology, Shanxi University, Taiyuan, China. His research interests include computer vision, pattern recognition, machine learning and human-computer interaction.
\end{IEEEbiography}
\begin{IEEEbiography}[{\includegraphics[width=1in,height=1.25in,clip,keepaspectratio]{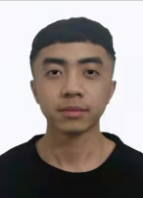}}]{Huijia Yan}
received the M.E. degrees in Computer technologies from Shanxi university, China, in 2025.

His research interests include computer vision, object detection, etc.
\end{IEEEbiography}
\begin{IEEEbiography}[{\includegraphics[width=1in,height=1.25in,clip,keepaspectratio]{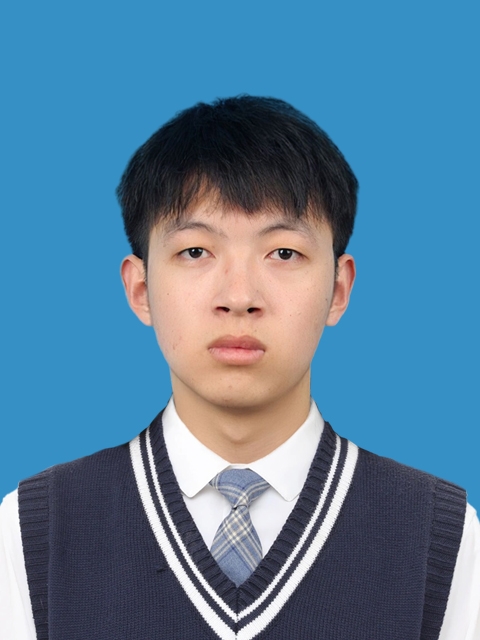}}]{Shaotong Qiao} received the B.S. degree in engineering from Shandong Jianzhu University, China, in 2024. He is currently pursuing the M.E. degrees in Computer technologies from Shanxi university, China.

His research interests include computer vision, person re-identification, etc.
\end{IEEEbiography}
\begin{IEEEbiography}[{\includegraphics[width=1in,height=1.25in,clip,keepaspectratio]{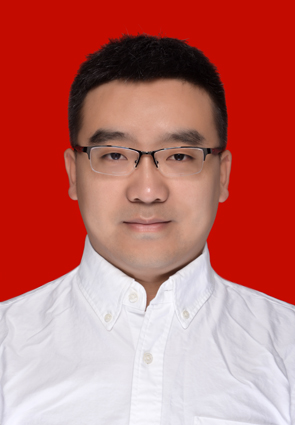}}]{Kai Feng}
received the Ph.D. degree in science from Shanxi University, China, in 2014. He is currently an Associate Professor at the School of Computer and Information Technology, Shanxi University, Taiyuan, China.

His research interests include combinatorial optimization and interconnection network analysis.
\end{IEEEbiography}

\begin{IEEEbiography}[{\includegraphics[width=1in,height=1.25in,clip,keepaspectratio]{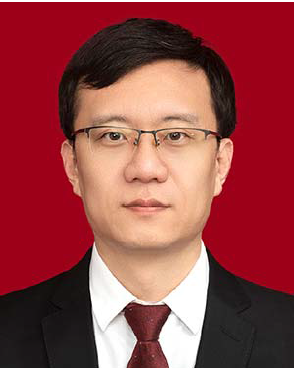}}]{Wei Wei}
(Member, IEEE) received the PhD degree in computer science from Shanxi University, in 2012. He is currently a full professor with the School of Computer and Information Technology, Shanxi University.

He has authored or coauthored more than 40 papers in his research fields, including the IEEE TPAMI, IEEE TKDE, NeurIPS, ICML, AAAI, IJCAI, and so on. His current research interests include data mining, machine learning, and embodied intelligence.

\end{IEEEbiography}
\end{document}